\documentclass[10pt,letterpaper,twocolumn]{article}
\usepackage[utf8]{inputenc}
\usepackage{color}
\usepackage{textcomp}
\usepackage{url}
\usepackage{tipa}
\usepackage{amsmath}
\usepackage{amssymb}
\usepackage{graphicx}
\usepackage{subscript}
\usepackage[unicode=true,
 bookmarks=false,
 breaklinks=true,pdfborder={0 0 1},backref=section,colorlinks=false]
 {hyperref}
\hypersetup{
 pagebackref=true,letterpaper=true,colorlinks}

\makeatletter

\pdfpageheight\paperheight
\pdfpagewidth\paperwidth

\newcommand{\lyxdot}{.}

\usepackage{cvpr}
\usepackage{times}
\usepackage{epsfig}
\usepackage{graphicx}

\cvprfinalcopy 

\def\cvprPaperID{3844} 

\ifcvprfinal\fi

\@ifundefined{showcaptionsetup}{}{%
 \PassOptionsToPackage{caption=false}{subfig}}
\usepackage{subfig}
\usepackage{authblk}
\makeatother

\begin{document}

\title{Priming Neural Networks}
\author[ ]{Amir Rosenfeld}
\author[ ]{Mahdi Biparva}
\author[ ]{John K.Tsotsos}
\affil[ ]{Department of Electrical Engineering and Computer Science\protect\\York University
\protect\\Toronto, ON, Canada, M3J 1P3}
\affil[ ]{\texttt {\{amir@eecs, mhdbprv@cse,tsotsos@cse\}.yorku.ca}}

\maketitle
\begin{abstract}
Visual priming is known to affect the human visual system to allow
detection of scene elements, even those that may have been near unnoticeable
before, such as the presence of camouflaged animals. This process
has been shown to be an effect of top-down signaling in the visual
system triggered by the said cue. In this paper, we propose a mechanism
to mimic the process of priming in the context of object detection
and segmentation. We view priming as having a modulatory, cue dependent
effect on layers of features within a network. Our results show how
such a process can be complementary to, and at times more effective
than simple post-processing applied to the output of the network,
notably so in cases where the object is hard to detect such as in
severe noise. Moreover, we find the effects of priming are sometimes
stronger when early visual layers are affected. Overall, our experiments
confirm that top-down signals can go a long way in improving object
detection and segmentation.
\end{abstract}

\section{Introduction}

\begin{figure}
\begin{centering}
\includegraphics[width=1\columnwidth]{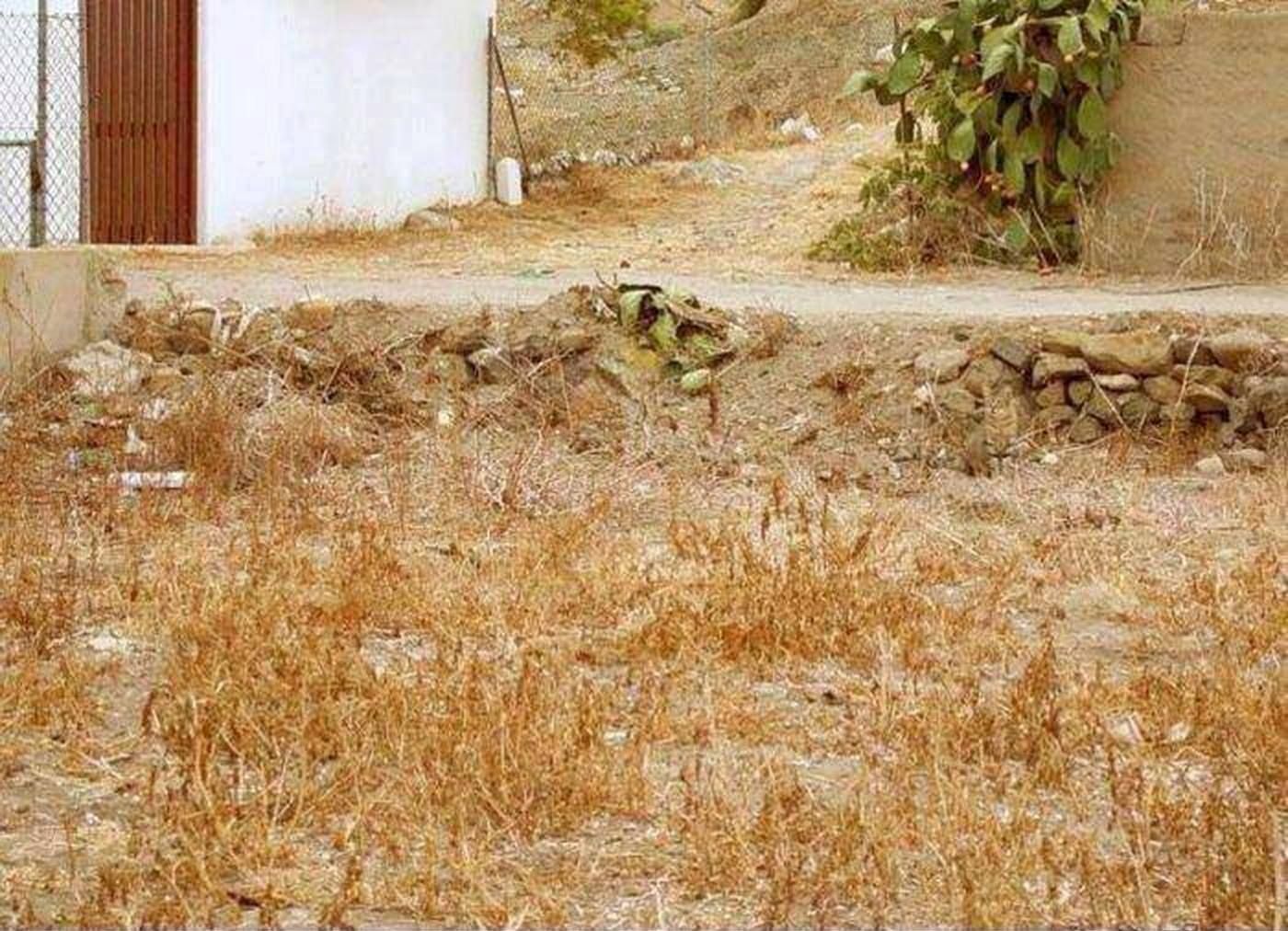}
\par\end{centering}
\caption{\label{fig:Visual-priming}Visual priming: something is hidden in
plain sight in this image. It is unlikely to notice it without a cue
on what it is (for an observer that has not seen this image before).
Once a cue is given, perception is modified to allow successful detection.
See footnote at bottom of this page for the cue, and supplementary
material for the full answer.}
\end{figure}

Psychophysical and neurophysiological studies of the human visual
system confirm the abundance of top-down effects that occur when an
image is observed. Such top-down signals can stem from either internal
(endogenous) processes of reasoning and attention or external (exogenous)
stimuli- i.e. cues - that affect perception (cf. \cite{tsotsos2011computational},
Chapter 3 for a more detailed breakdown). External stimuli having
such effects are said to \emph{prime} the visual system, and potentially
have a profound effect on an observer's perception. This often results
in an ``Aha!'' moment for the viewer, as he/she suddenly perceives
the image differently; Fig. \ref{fig:Visual-priming} shows an example
of such a case. We make here the distinction between 3 detection strategies:
(1) \emph{free viewing}, (2) \emph{priming }and (3) \emph{pruning}.
Freely viewing the image, the default strategy, likely reveals nothing
more than a dry grassy field near a house. Introducing a cue about
a target in the image\footnote{Object in image: \textturnt \textturna \textopeno{}}
results in one of two possibilities. The first, also known as priming,
is modification to the computation performed when viewing the scene
with the cue in mind. The second, which we call pruning - is a modification
to the decision process after all the computation is finished. When
the task is to detect objects, this can mean retaining all detections
match the cue, even very low confidence ones and discarding all others.
While both are viable ways to incorporate the knowledge brought on
by the cue, priming often highly increases the chance of detecting
the cued object. Viewing the image for an unlimited amount of time
and pruning the results is less effective; in some cases, detection
is facilitated only by the cue. We claim that priming allows the cue
to affect the visual process from early layers, allowing detection
where it was previously unlikely to occur in free-viewing conditions.
This has also recently gained some neurophysiological evidence \cite{bartsch2017attention}.

In this paper, we propose a mechanism to mimic the process of visual
priming in deep neural networks in the context of object detection
and segmentation. The mechanism transforms an external cue about the
presence of a certain class in an image (e.g., ``person'') to a
modulatory signal that affects all layers of the network. This modulatory
effect is shown via experimentation to significantly improve object
detection performance when the cue is present, more so than a baseline
which simply applies post-processing to the network's result. Furthermore,
we show that priming early visual layers has a greater effect that
doing so for deeper layers. Moreover, the effects of priming are shown
to be much more pronounced in difficult images such as very noisy
ones. 

The remainder of the paper is organized as follows: in Sec. \ref{sec:Related-Work}
we go over related work from computers vision, psychology and neurophysiology.
In Sec. \ref{sec:Approach} we go over the details of the proposed
method. In Sec. \ref{sec:Experiments} we elaborate on various experiments
where we evaluate the proposed method in scenarios of object detection
and segmentation. We finish with some concluding remarks. 

\section{Related Work\label{sec:Related-Work}}

\begin{figure}
\includegraphics[width=1\columnwidth]{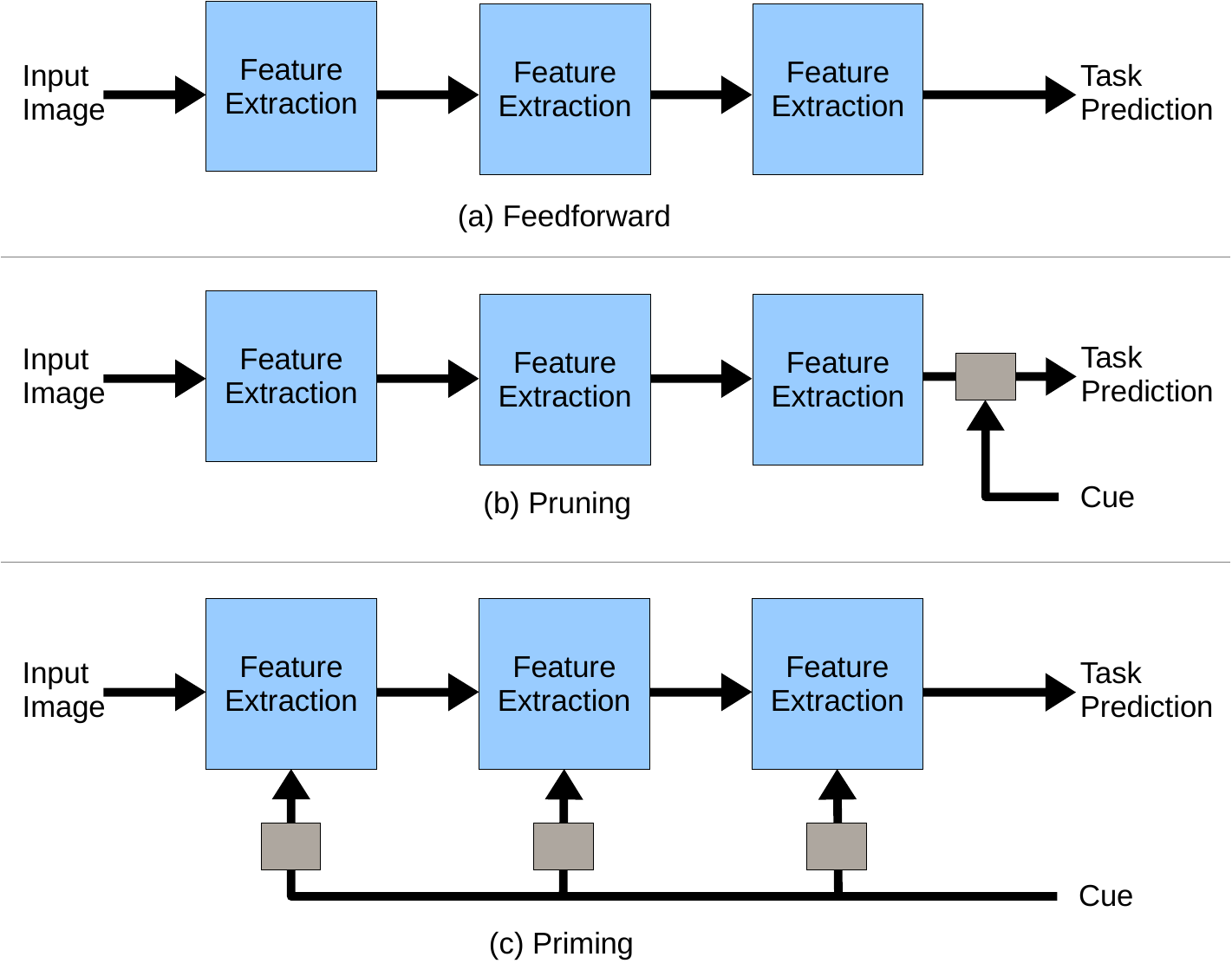}

\caption{\label{fig:comparison}A neural network can be applied to an input
in an either unmodified manner (\emph{top}), pruning the results after
running (\emph{middle}) or \emph{priming }the network via an external
signal (cue) in image to affect all layers of processing (\emph{bottom}).}
\end{figure}

Context has been very broadly studied in cognitive neuroscience \cite{biederman1982scene,biederman1981semantics,oliva2007role,tulving1990priming,wig2005reductions,palmer1975effects,hollingworth1998does}
and in computer vision \cite{galleguillos2010context,divvala2009empirical,torralba2003context,torralba2003contextual,rabinovich2007objects,yao2012describing,murphy2004using}.
It is widely agreed \cite{shrivastava2016contextual} that context
plays crucial role for various visual tasks. Attempts have been made
to express a tangible definition for context due to the increased
use in the computer vision community \cite{torralba2003context,torralba2003contextual}
. 

Biederman et al. \cite{biederman1982scene} hypothesizes object-environments
dependencies into five categories: probability, interposition, support,
familiar size, position. Combinations of some of these categories
would form a source of contextual information for tasks such as object
detection \cite{torralba2003contextual,shrivastava2016contextual},
semantic segmentation \cite{harley2017segmentation}, and pose estimation
\cite{carreira2016human}. Context consequently is the set of sources
that partially or collectively influence the perception of a scene
or the objects within \cite{strat1993employing}. 

Visual cues originated from contextual sources, depending on the scope
they influence, further direct visual tasks at either global or local
level \cite{torralba2003context,torralba2003contextual}. Global context
such as scene configuration, imaging conditions, and temporal continuity
refers to cues abstracted across the whole scene. On the other hand,
local context such as semantic relationships and local-surroundings
characterize associations among various parts of similar scenes.

Having delineated various contextual sources, the general process
by which the visual hierarchy is modulated prior to a particular task
is referred to as visual priming \cite{tsotsos2011computational,posner1978attended}.
A cue could be provided either implicitly by a contextual source or
explicitly through other modalities such as language.

There has been a tremendous amount of work on using some form of top-down
feedback to contextually prime the underlying visual representation
for various tasks \cite{tulving1990priming,wig2005reductions,palmer1975effects,hollingworth1998does}.
The objective is to have signals generated from some task such that
they could prepare the visual hierarchy oriented for the primary task.
\cite{shrivastava2016contextual} proposes contextual priming and
feedback for object detection using the Faster R-CNN framework \cite{ren2015faster}.
The intuition is to modify the detection framework to be able to generate
semantic segmentation predictions in one stage. In the second stage,
the segmentation primes both the object proposal and classification
modules. 

Instead of relying on the same modality for the source of priming,
\cite{de2017modulating,perez2017film} proposes to modulate features
of a visual hierarchy using the embeddings of the language model trained
on the task of visual question answering \cite{antol2015vqa,johnson2016clevr}.
In other words, using feature-wise affine transformations, \cite{perez2017film}
multiplicatively and additively modulates hidden activities of the
visual hierarchy using the top-down priming signals generated from
the language model, while \cite{shrivastava2016contextual} append
directly the semantic segmentation predictions to the visual hierarchy.
Recently, \cite{harley2017segmentation} proposes to modulate convolutional
weight parameters of a neural networks using segmentation-aware masks.
In this regime, the weight parameters of the model are directly approached
for the purpose of priming.

Although all these methods modulate the visual representation, none
has specifically studied the explicit role of category cues to prime
the visual hierarchy for object detection and segmentation. In this
work, we strive to introduce a consistent parametric mechanism into
the neural network framework. The proposed method allows every portion
of the visual hierarchy to be primed for tasks such as object detection
and semantic segmentation. It should be noted that this use of priming
was defined as part of the Selective Tuning (ST) model of visual attention
\cite{tsotsos1995SelTun}. Other aspects of ST have recently appeared
as part of classification and localization networks as well \cite{biparva2017stnet,zhang2016top},
and our work explores yet another dimension of the ST theory.

\section{Approach\label{sec:Approach}}

Assume that we have some network $N$ to perform a task such as object
detection or segmentation on an image $I$. In addition, we are given
some cue $h\in\mathcal{R^{\text{n}}}$ about the content of the image.
We next describe pruning and priming, how they are applied and how
priming is learned. We assume that $h$ is a binary encoding of them
presence of some target(s) (e.g, objects) - though this can be generalized
to other types of information. For instance, an explicit specification
of color, location, orientation, etc, or an encoded features representation
as can be produced by a vision or language model. Essentially, one
can either ignore this cue, use it to post-process the results, or
use it to affect the computation. These three strategies are presented
graphically in Fig. \ref{fig:comparison}.

\subparagraph{Pruning.\label{par:Pruning}}

In pruning, $N$ is fed an image and we use $h$ to post-process the
result. In object detection, all bounding boxes output by $N$ whose
class is different than indicated by $h$ are discarded. For segmentation,
assume $N$ outputs a score map of size $C\times h\times w$ , where
$L$ is the number of classes learned by the network, including a
background class. We propose two methods of pruning, with complementary
effects. The first type increases recall by ranking the target class
higher: for each pixel (x,y), we set the value of all score maps inconsistent
with $h$ to be $-\infty$ , except that of the background. This allows
whatever detection of the hinted class to be ranked higher than other
which previously masked it. The second type simply sets each pixels
which was not assigned by the segmentation the target class to the
background class. This decreases recall but increases the precision.
These types of pruning are demonstrated in Fig. \ref{fig:summary-segmentation}
and discussed below.

\paragraph{Priming.}

Our approach is applicable to any network $N$ with a convolutional
structure, such as a modern network for object detection, e.g. \cite{liu2016ssd}.
To enable priming, we freeze all weights in $N$ and add a parallel
branch $N_{p}$. The role of $N_{p}$ is to transform an external
cue $h\in\mathcal{R^{\text{n}}}$ to modulatory signals which affect
all or some of the layers of $N$. Namely, let $L_{i}$ be some layer
of $N.$ Denote the output of $L_{i}$ by $x_{i}\in\mathcal{R}^{c_{i}\times h_{i}\times w_{i}}$
where $c_{i}$ is the number of feature planes and $h_{i},w_{i}$
are the height and width of the feature planes. Denote the $j_{th}$
feature plane of $x_{i}$ by $x_{ij}\in\mathcal{R}^{h_{i}\times w_{i}}$. 

$N_{p}$ modulates each feature plane $x_{ij}$ by applying to the

\begin{equation}
f_{ij}(x_{ij},h)=\hat{x}_{ij}
\end{equation}

The function $f_{ij}$ always operates in a spatially-invariant manner
- for each element in a feature plane, the same function is applied.
Specifically, we use a simple residual function, that is 

\begin{equation}
\hat{x}_{ij}=\alpha_{ij}\cdot x_{ij}+x_{ij}\label{eq:modulation}
\end{equation}

Where the coefficients $\boldsymbol{\alpha_{i}=}[\alpha_{i1},\dots,\alpha_{ic_{i}}]^{T}$
are determined by a linear transformation of the cue:

\begin{equation}
\alpha_{i}=W_{i}*h
\end{equation}

\begin{figure*}
\includegraphics[width=1\textwidth]{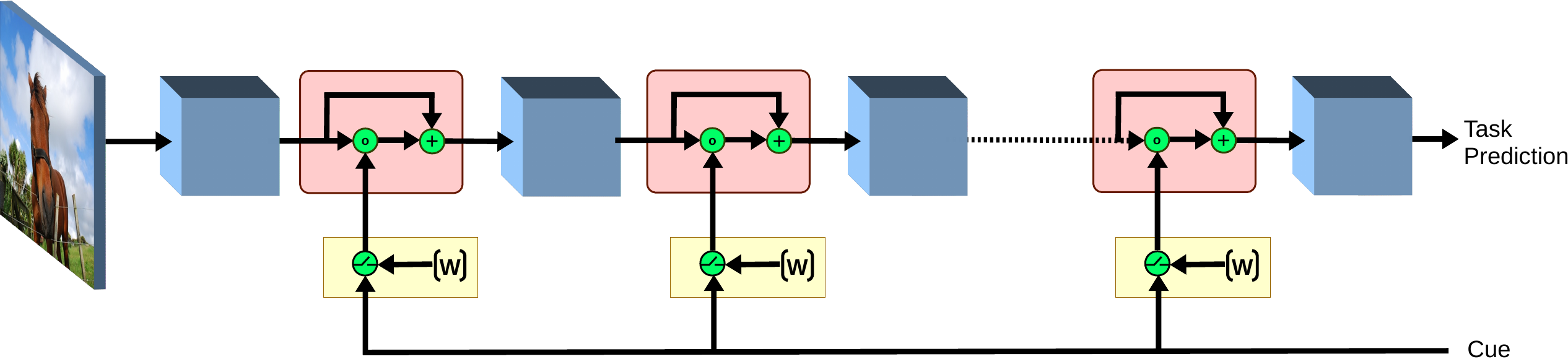}

\caption{\label{fig:Overall-view-of}Overall view of the proposed method to
prime deep neural networks. A cue about some target in the image is
given by and external source or some form of feedback. The process
of priming involves affecting each layer of computation of the network
by modulating representations along the path. }
\end{figure*}

An overall view of the proposed method is presented in Fig. \ref{fig:Overall-view-of}.

\subparagraph{Types of Modulation}

The modulation in eq. \ref{eq:modulation} simply adds a calculated
value to the feature plane. We have experimented with other types
of modulation, namely non-residual ones (e.g, purely multiplicative),
as well as following the modulated features with a non-linearity (ReLU),
or adding a bias term in addition to the multiplicative part. The
single most important dominant ingredient to reach good performance
was the residual formulation - without it, training converged to very
poor results. The formulation in eq. \ref{eq:modulation} performed
best without any of the above listed modifications. We note that an
additive model, while having converged to better results, is not fully
consistent with biologically plausible models (\cite{tsotsos1995SelTun})
which involve suppression/selection of visual features, however, it
may be considered a first approximation.

\paragraph{Types of Cues}

The simplest form of a cue $h$ is an indicator vector of the object(s)
to be detected, i.e, a vector of 20 zeros and 1 in the coordinate
corresponding to ``horse'', assuming there are 20 possible object
classes, such as in Pascal  \cite{everingham2010pascal}. We call
this a \emph{categorical }cue\textbf{ }because it explicitly carries
semantic information about the object. This means that when a single
class $k$ is indicated, $\alpha_{i}$ becomes the $k_{th}$ column
of $W_{i}$. 

\subsection{Training}

To learn how to utilize the cue, we freeze the parameters of our original
network $N$ and add the network block $N_{p}$. During training,
with each training example ($I_{i},y_{i}$) fed into $N$ we feed
$h_{i}$ into $N_{p}$, where $I_{i}$ is an image, $y_{i}$ is the
ground-truth set of bounding boxes and $h_{i}$ is the corresponding
cue. The output and loss functions of the detection network remain
the same, and the error is propagated through the parameters of $N_{p}$.
Fig. \ref{fig:Overall-view-of} illustrates the network. $N_{p}$
is very lightweight with respect to $N$, as it only contains parameters
to transform from the size of the cue $h$ to at most $K=\sum_{i}k_{i}$
where $k_{i}$ is the number of output feature planes in each layer
of the network. 

\subparagraph{Multiple Cues Per Image.}

Contemporary object detection and segmentation benchmarks \cite{lin2014microsoft,everingham2010pascal}
often contain more than one object type per image. In this case, we
may set each coordinate in $h$ to 1 iff the corresponding class is
present in the image. However, this tends to prevent $N_{p}$ from
learning to modulate the representation of $N$ in a way which allows
it to suppress irrelevant objects. Instead, if an image contains $k$
distinct object classes, we duplicate the training sample $k$ times
and for each duplicate set the ground truth to contain only one of
the classes. This comes at the expense of a longer training time,
depending on the average number $k$ over the dataset. 

\section{Experiments\label{sec:Experiments}}

We evaluate our method on two tasks: object detection and object class
segmentation. In each case, we take a pre-trained deep neural network
and explore how it is affected by priming or pruning. Our goal here
is not necessarily to improve state-of-the-art results but rather
to show how usage of top-down cues can enhance performance. Our setting
is therefore different than standard object-detection/segmentation
scenarios: we assume that some cue about the objects in the scene
is given to the network and the goal is to find how it can be utilized
optimally. Such information can be either deduced from the scene,
such as in contextual priming \cite{shrivastava2016contextual,katti2016object}
or given by an external source, or even be inferred from the task,
such as in question answering \cite{antol2015vqa,johnson2016clevr}. 

Our experiments are conducted on the Pascal VOC \cite{everingham2010pascal}
2007 and 2012 datasets. For priming object detection networks we use
pre-trained models of SSD \cite{liu2016ssd} and yolo-v2 \cite{redmon2016yolo9000}
and for segmentation we use the FCN-8 segmentation network of \cite{long2015fully}
and the DeepLab network of \cite{chen2016deeplab}. We use the YellowFin
optimizer \cite{zhang2017yellowfin} in all of our experiments, with
a learning rate of either 0.1 or 0.01 (depending on the task). 

\subsection{Object Detection\label{subsec:Object-Detection}}

\begin{figure}
\centering{}\subfloat[]{\begin{raggedright}
\includegraphics[height=0.2\columnwidth]{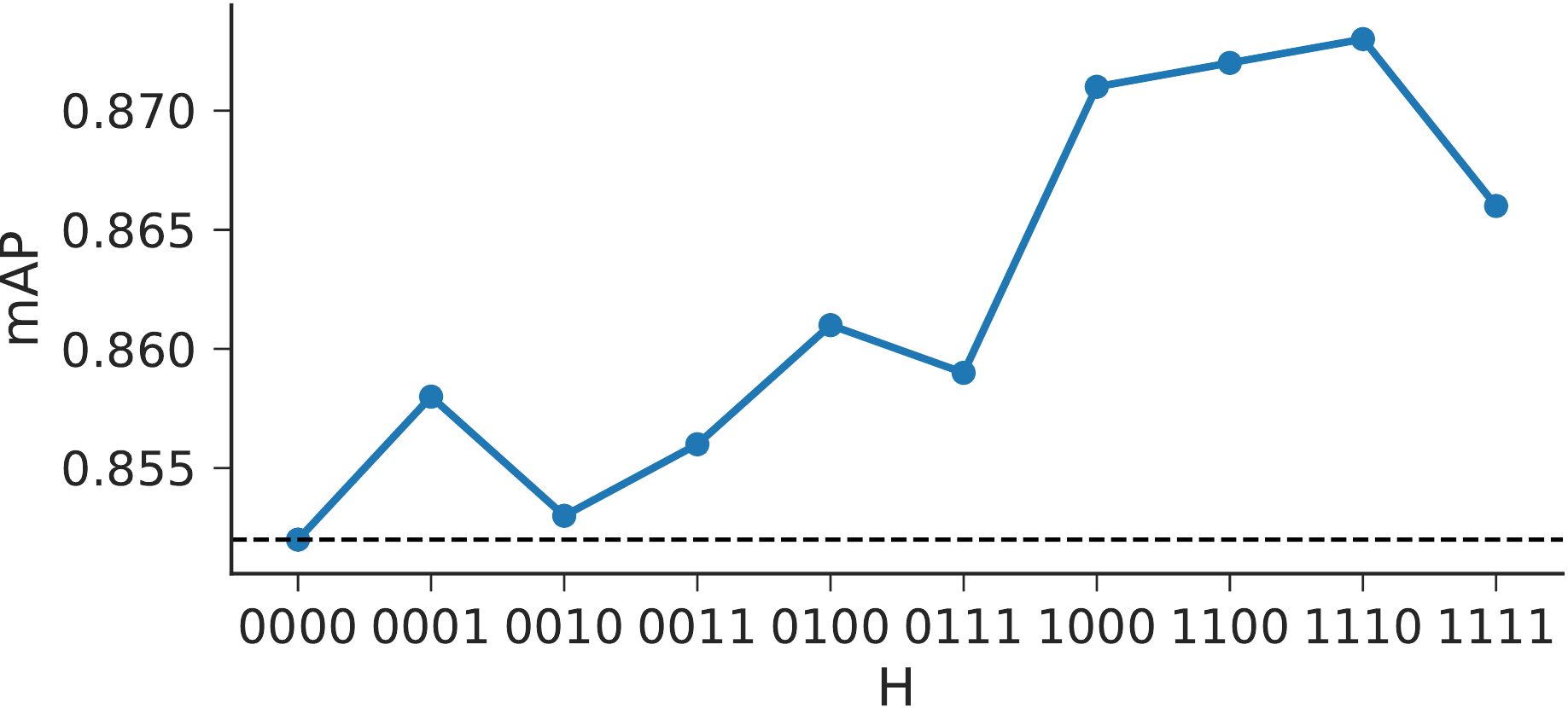}
\par\end{raggedright}
}\subfloat[]{\includegraphics[height=0.2\columnwidth]{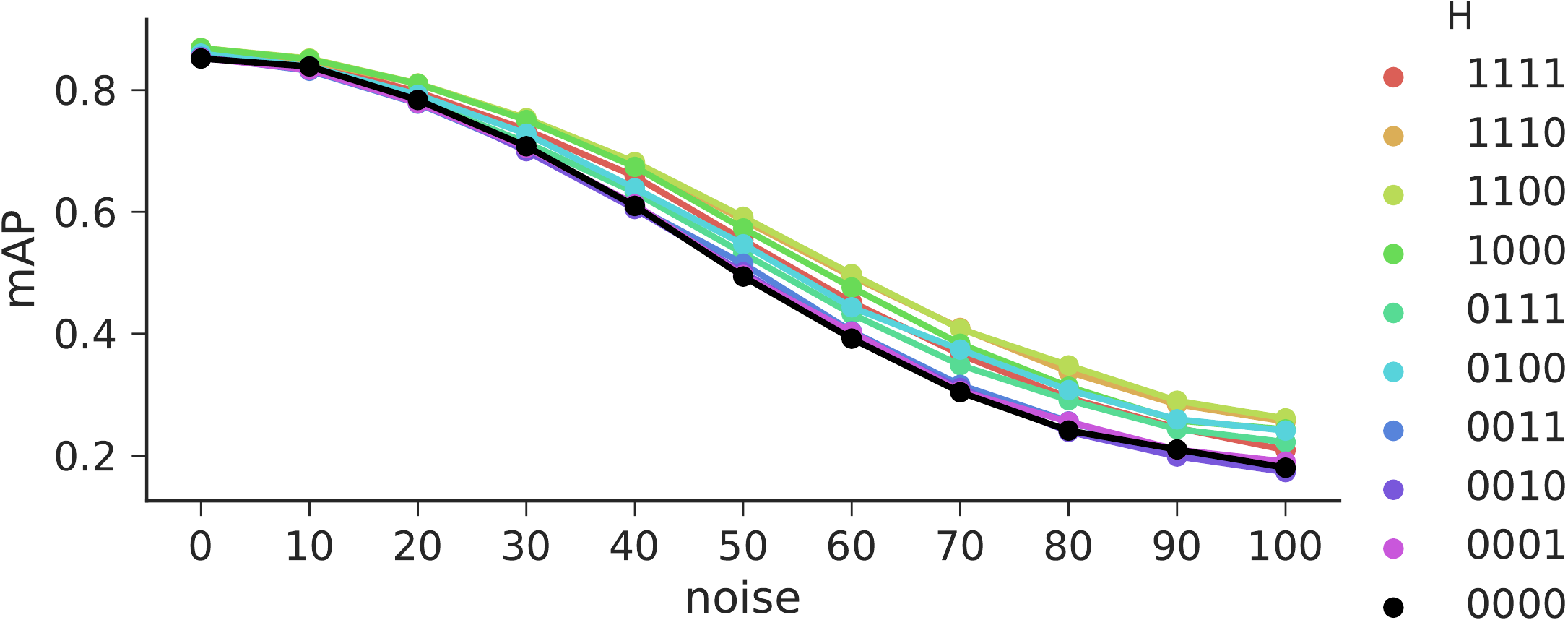}

}\caption{(a) \label{fig:h-and-noise}Performance gains by priming different
parts of the SSD objects detector. Priming early parts of the network
causes the most significant boost in performance. black dashed line
shows performance by pruning. (b) Testing variants of priming against
increasing image noise. The benefits of priming become more apparent
in difficult viewing conditions. The x axis indicates which block
of the network was primed (1 for primed, 0 for not primed). }
\end{figure}

We begin by testing our method on object detection. Using an implementation
of SSD \cite{liu2016ssd}, we apply a pre-trained detector trained
on the trainval sets of Pascal 2012+2007 to the test set of Pascal
2007. We use the SSD-300 variant as described in the paper. In this
experiment, we trained and tested on what we cal PAS\textsuperscript{\#}:
this is a reduced version of Pascal-2007 containing only images with
a single object class (but possibly multiple instances). We use this
reduced dataset to test various aspects of our method, as detailed
in the following subsections. Without modification, the detector attains
a mAP (mean-average precision) of 81.4\% on PAS\textsuperscript{\#}(77.4\%
on the full test set of Pascal 2007). Using simple pruning as described
above, this increases to 85.2\%. This large boost in performance is
perhaps not surprising, since pruning effectively removes all detections
of classes that do not appear in the image. The remaining errors are
those of false alarms of the ``correct'' class or mis-detections. 

\subsubsection{Deep vs Shallow Priming}

We proceed to the main result, that is, how priming affects detection.
The SSD object detector contains four major components: (1) a pre-trained
part made up of some of the layers of vgg-16 \cite{simonyan2014very}
(a.k.a the ``base network'' in the SSD paper), (2) some extra convolutional
layers on top of the vgg-part, (3) a localization part and (4) a class
confidence part. We name these part \emph{vgg}, \emph{extra}, \emph{loc
}and \emph{conf }respectively. 

To check where priming has the most significant impact, we select
different subsets of these components and denote them by 4-bit binary
vectors $s_{i}\in\{0,1\}^{4}$, where the bits correspond from left
to right to the vgg,extra,localization and confidence parts. For example,
$s=1000$ means letting $N_{p}$ affect only the earliest (\emph{vgg})
part of the detector, while all other parts remain unchanged by the
priming (except indirectly affecting the deeper parts of the net).
We train $N_{p}$ on 10 different configurations: these include priming
from the deepest layers to the earliest: $1111$, $0111$, $0011$,
$0001$ and from the earliest layer to the deepest: $1000$, $1100$,
$1110$. We add $0100$ and $0010$ to check the effect of exclusive
control over middle layers and finally $0000$ as the default configuration
in which $N_{p}$ is degenerate and the result is identical to pruning.
Fig \ref{fig:h-and-noise} (a) shows the effect of priming each of
these subsets of layers on PAS\textsuperscript{\#}. Priming early
layers (those at the bottom of the network) has a much more pronounced
effect than priming deep layers. The single largest gain by priming
a single component is for the \emph{vgg} part: $1000$ boosts performance
from 85\% to 87.1\%. A smaller gain is attained by the \emph{extra}
component: 86.1\% for $0100$. The performance peaks at \textbf{87.3}\%
for 1110, though this is only marginally higher than attained by $1100$
- priming only the first two parts. 

\subsubsection{Ablation Study}

\begin{figure}
\includegraphics[width=1\columnwidth]{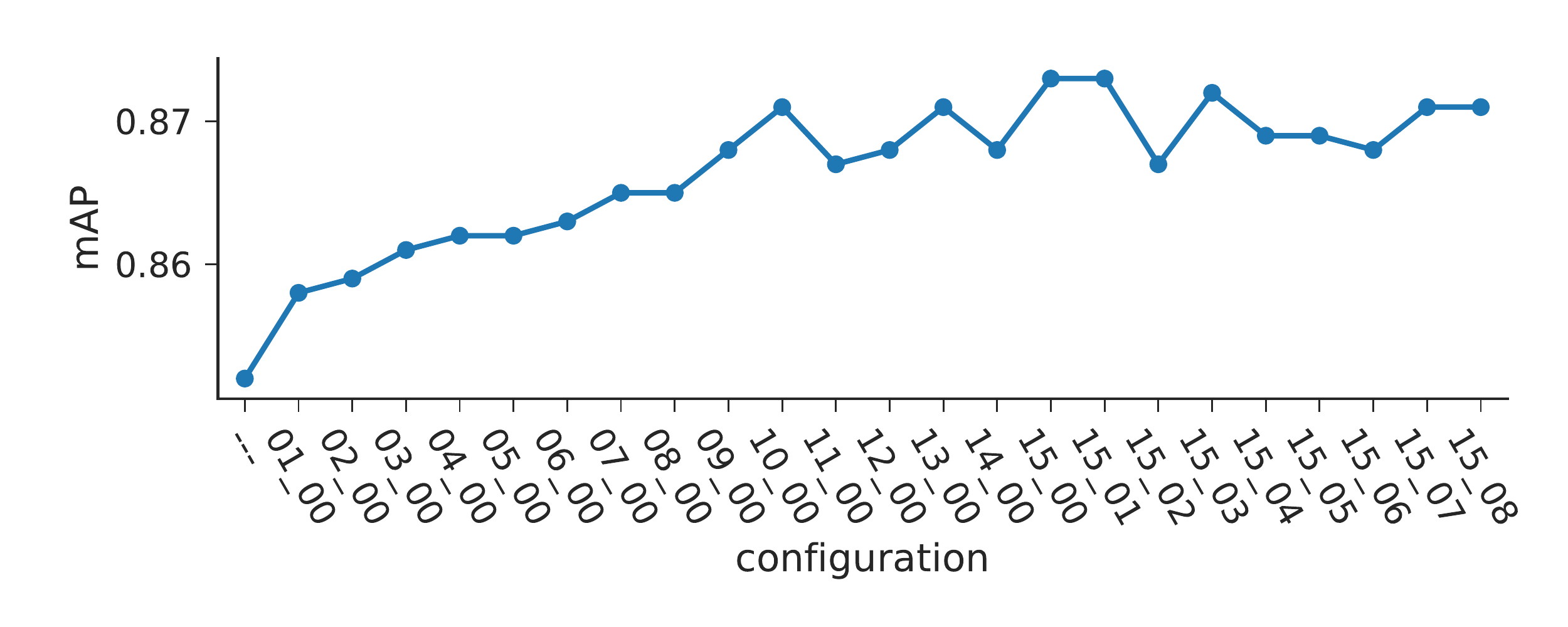}

\caption{\label{fig:early-ablation}Effects of early priming: we show how mAP
increases when we allow priming to affect each time another layer,
from the very bottom of the network. Priming early layers has a more
significant effect than doing so for deeper ones. The numbers indicate
how many layers were primed from the first,second blocks of the SSD
network, respectively. }
\end{figure}

Priming the earliest layers (\emph{vgg}+\emph{extra}) of the SSD object
detector brings the most significant boost in performance. The first
component described above contains 15 convolutional layers and the
second contains 8 layers, an overall total of 23. To see how much
we can gain with priming on the first few layers, we checked the performance
on PAS\textsuperscript{\#} when training on the first $k$ layers
only, for each $k\in\{1,2,\dots23\}$. Each configuration was trained
for 4000 iterations. Fig. \ref{fig:early-ablation} shows the performance
obtained by each of these configurations, where $i$\_$j$ in the
x-axis refers to having trained the first $i$ layers and the first
$j$ layers of the first and second parts respectively. We see that
the very first convolutional layer already boosts performance when
primed. The improvement continues steadily as we add more layers and
fluctuates around 87\% after the 15th layer. The fluctuation is likely
due to randomness in the training process. This further shows that
priming has strong effects when applied to very early layers of the
network. 
\begin{figure*}
\begin{centering}
\subfloat[]{\includegraphics[width=1\columnwidth]{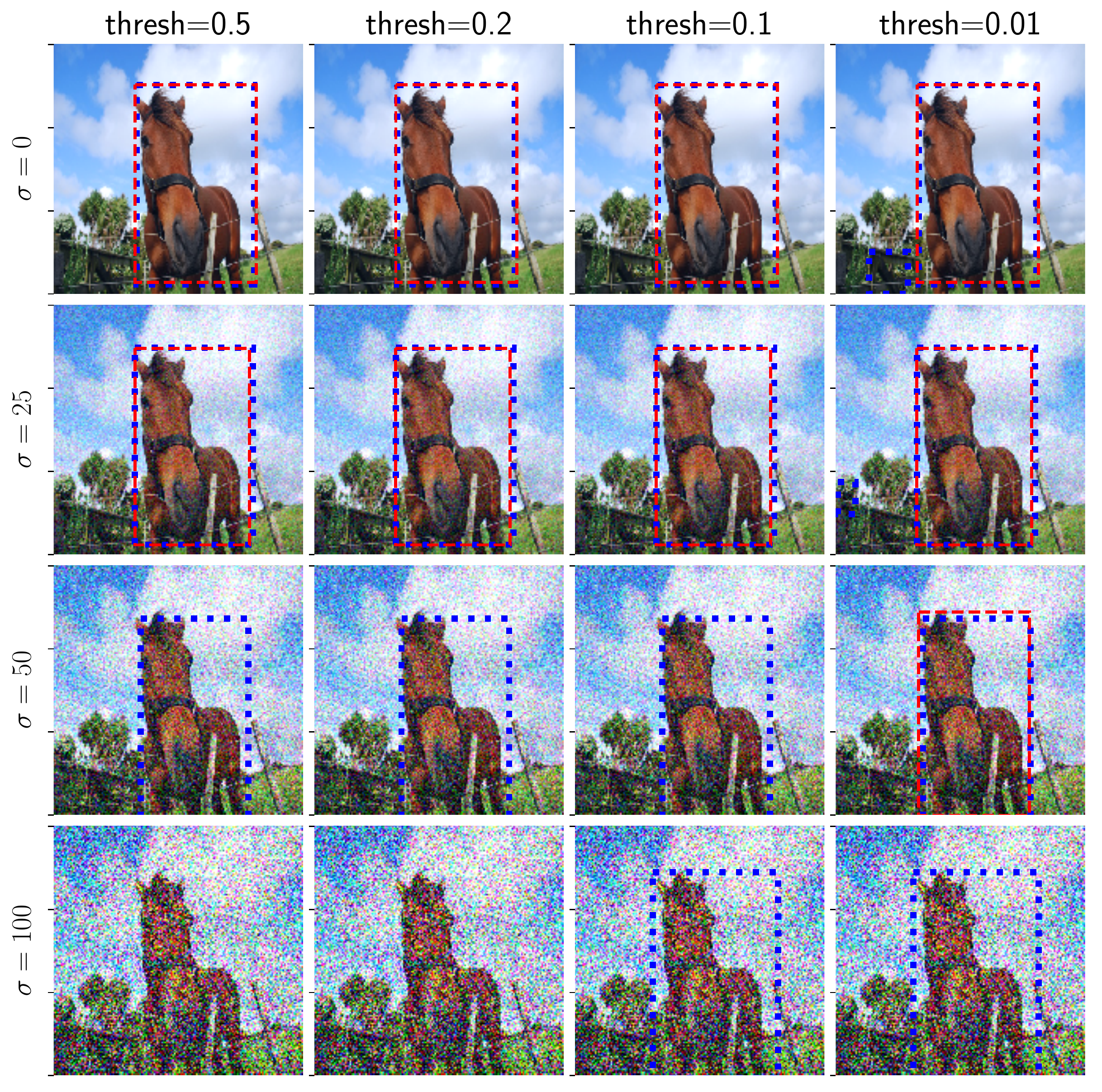}}\subfloat[]{\includegraphics[width=1\columnwidth]{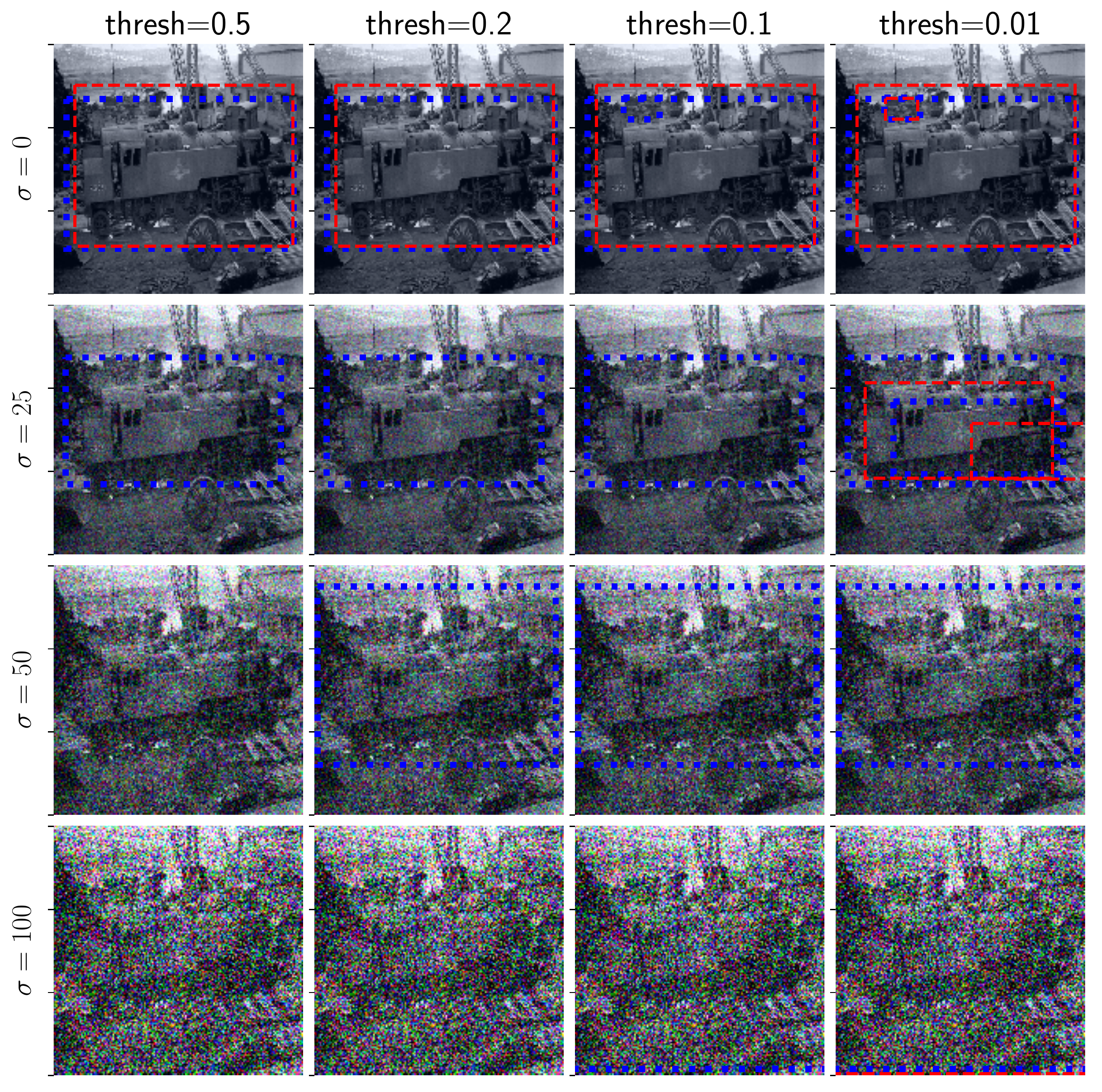}}
\par\end{centering}
\caption{\label{fig:Priming-vs.-Pruning.}Priming vs. Pruning. Priming a detector
allows it to find objects in images with high levels of noise while
mostly avoiding false-alarms. Left to right (\emph{a,b}): decreasing
detection thresholds (increasing sensitivity). Top to bottom: increasing
levels of noise. Priming (\textbf{\textcolor{blue}{blue}} dashed boxes)
is able to detect the horse (\emph{a}) across all levels of noise,
while pruning (\textbf{\textcolor{red}{red }}dashed boxes) does not.
For the highest noise level, the original classifier does not detect
the horse at all - so pruning is ineffective. (\emph{b}) Priming enables
detection of the train for all but the most severe level of noise.
Decreasing the threshold for pruning only produces false alarms. We
recommend viewing this figure in color on-line.}
\end{figure*}

\subsubsection{Detection in Challenging Images}

As implied by the introduction, perhaps one of the cases where the
effect of priming is stronger is when facing a challenging image,
such as adverse imaging conditions, low lighting, camouflage, noise.
As one way to test this, we compared how priming performs under noise.
We took each image in the test set of Pascal 2007 and added random
Gaussian noise chosen from a range of standard deviations, from 0
to 100 in increments of 10. The noisy test set of PAS\textsuperscript{\#}
with variance $\sigma$ is denoted PAS\textsuperscript{\#}\textsubscript{N($\sigma$)}.
For each $\sigma$, we measure the mAP score attained by either pruning
or priming. Note that none of our experiments involved training with
only images - these are only used for testing. We plot the results
in Fig. \ref{fig:h-and-noise} (b). As expected, both methods suffer
from decreasing accuracy as the noise increases. However, priming
is more robust to increasing levels of noise; the difference between
the two methods peaks at a moderate level of noise, that is, $\sigma=80$,
with an advantage of $\boldsymbol{10.7}$\% in mAP: 34.8\% compared
to 24.1\% by pruning. The gap decreases gradually to 6.1\% (26.1\%
vs 20\%) for a noise level of $\sigma=100$. We believe that this
is due to the early-layer effects of priming on the network, selecting
features from the bottom up to match the cue. Fig \ref{fig:Priming-vs.-Pruning.}
shows qualitative examples, comparing priming versus pruning: we increase
the noise from top to bottom and decrease the threshold (increase
the sensitivity) from left to right. We show in each image only the
top few detections of each method to avoid clutter. Priming allows
the detector to find objects in images with high levels of noise (see
lower rows of a,b). In some cases priming proves to be \emph{essential
}for the detection: lowering the un-primed detector's threshold to
a minimal level does not increase the recall of the desired object
(a, 4th row); in fact, it only increases the number of false alarms
(b, 2nd row, last column). Priming, on the other hand, is often less
sensitive to a low threshold and the resulting detection persists
along a range thereof. 

\subsection{Cue Aware Training}

In this section, we also test priming on an object detection task
as well as segmentation with an added ingredient - multi-cue training
and testing. In Sec. \ref{subsec:Object-Detection} we limited ourselves
to the case where there is only one object class per image. This limitation
is often unrealistic. To allow multiple priming cues per image, we
modify the training process as follows: for each training sample $<I,gt>$
containing object classes $c_{1},\dots c_{k}$ we split the training
example for $I$ to $k$ different tuples $<I_{i},h_{i},gt_{i}>,i\in\{1\dots k\}$,
where $I_{i}$ are all identical to $I$, $h_{i}$ indicate the presence
of class $c_{i}$ and $gt_{i}$ is the ground-truth $gt$ reduced
to contain only the objects of class $c_{i}$ - meaning the bounding
boxes for detection, or the masks for segmentation. This explicitly
coerces the priming network $N_{p}$ to learn how to force the output
to correspond to the given cue, as the input image remains the same
but the cue and desired output change together. We refer to this method
multi-cue aware training (CAT for short) , and refer to the unchanged
training scheme as regular training. 

\subsubsection{Multi-Cue Segmentation }

Here, we test the multi-cue training method on object class segmentation.
We begin with the FCN-8 segmentation network of \cite{long2015fully}.
We train on the training split of SBD (Berkeley Semantic Boundaries
Dataset and Benchmark) dataset \cite{hariharan2011semantic}, as is
done in \cite{zhao2016pyramid,chen2016deeplab,dai2015boxsup,long2015fully}.
We base our code on an unofficial PyTorch\footnote{\url{http://pytorch.org/}}
implementation\footnote{\url{https://github.com/wkentaro/pytorch-fcn}}.
Testing is done of the validation set of Pascal 2011, taking care
to avoid overlapping images between the training set defined by \cite{hariharan2011semantic}
\footnote{for details, please refer to \url{https://github.com/shelhamer/fcn.berkeleyvision.org/tree/master/data/pascal}},
which leaves us with 736 validation images. The baseline results average
IOU score of 65.3\%. As before, we let the cue be a binary encoding
of the classes present in the image. We train and test the network
in two different modes: one is by setting for each training sample
(and testing) the cue so $h_{i}=1$ if the current image contains
at least one instance of class $i$ and 0 otherwise. The other is
the multi-cue method we describe earlier, i.e , splitting each sample
to several cues with corresponding ground-truths so each cue is a
one-hot encoding, indicating only a single class. For both training
strategies, testing the network with a cue creates a similar improvement
in performance, from 65.3\% to 69\% for regular training and to 69.2\%
for multi-cue training. 

\begin{figure}
\begin{centering}
\includegraphics[width=0.24\columnwidth]{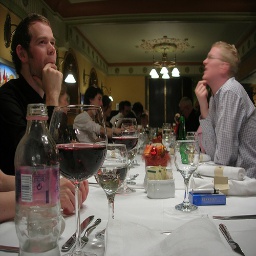}\,\includegraphics[width=0.24\columnwidth]{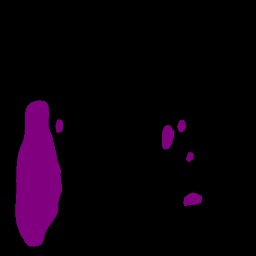}\,\includegraphics[width=0.24\columnwidth]{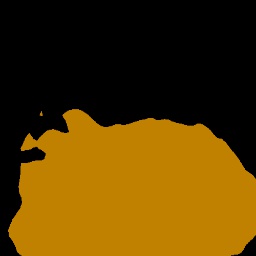}\,\includegraphics[width=0.24\columnwidth]{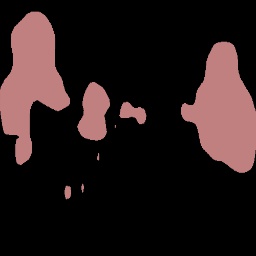}
\par\end{centering}
\begin{centering}
\includegraphics[width=0.24\columnwidth]{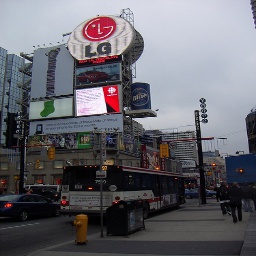}\,\includegraphics[width=0.24\columnwidth]{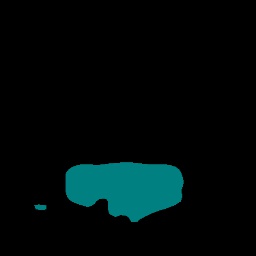}\,\includegraphics[width=0.24\columnwidth]{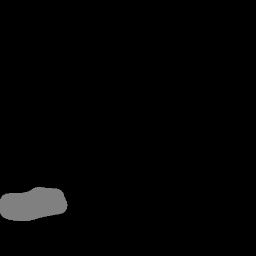}\,\includegraphics[width=0.24\columnwidth]{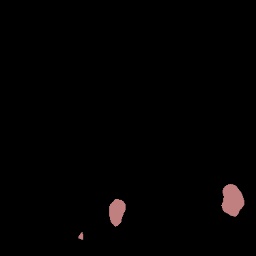}
\par\end{centering}
\caption{\label{fig:Effect-of-training}Effect of priming a segmentation network
with different cues. In each row, we see an input image and the output
of the network when given different cues. Top row: cues are resp.
bottle, diningtable, person. Given a cue (e.g, \emph{bottle}), the
network becomes more sensitive to bottle-like image structures while
suppressing others. This happens not by discarding results but rather
by affecting computation starting from the early layers. }
\end{figure}

The main advantage of the multi-cue training is that it allows the
priming network $N_{p}$ to force $N$ to focus on different objects
in the image. This is illustrated in Fig. \ref{fig:Effect-of-training}.
The top row of the figure shows from left to right an input image
and the resulting segmentation masks when the network is cued with
classes \emph{bottle}, \emph{diningtable }and \emph{person}. The bottom
row is cued with \emph{bus}, \emph{car}, \emph{person}. The cue-aware
training allows the priming network to learn how to suppress signals
relating to irrelevant classes while retaining the correct class from
the bottom-up. 

\subparagraph{Types of Pruning.\label{par:Types-of-Pruning.}}

\begin{figure}
\begin{centering}
\subfloat[input]{\includegraphics[width=0.155\columnwidth]{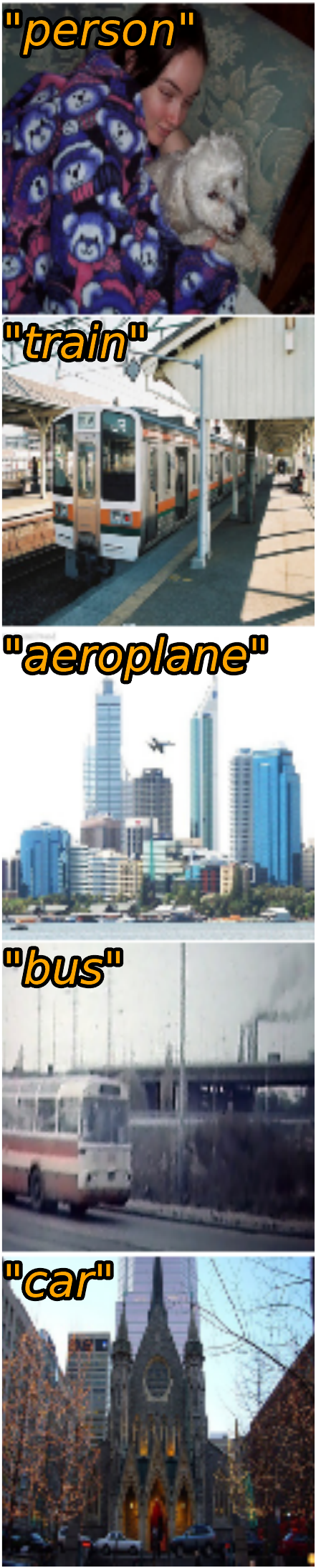}

}\subfloat[gt]{\includegraphics[width=0.155\columnwidth]{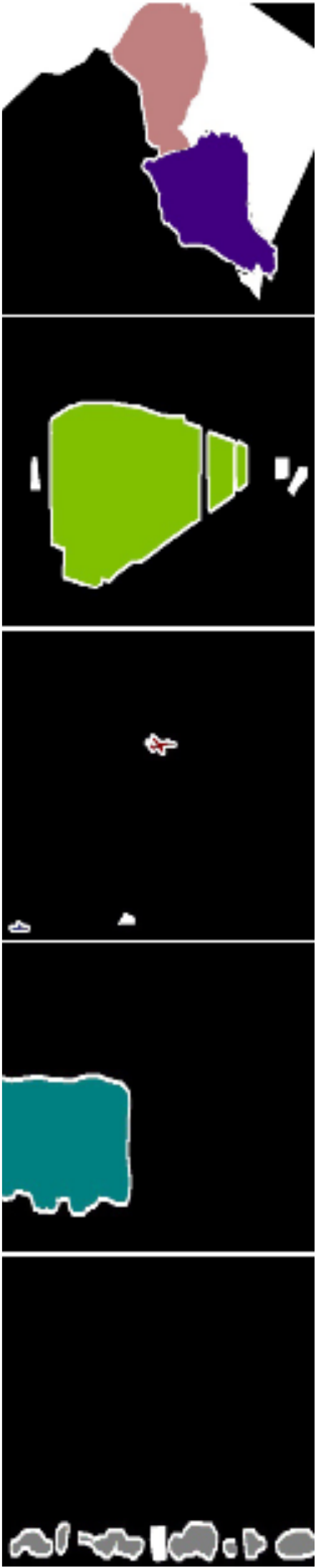}

}\subfloat[regular]{\includegraphics[width=0.155\columnwidth]{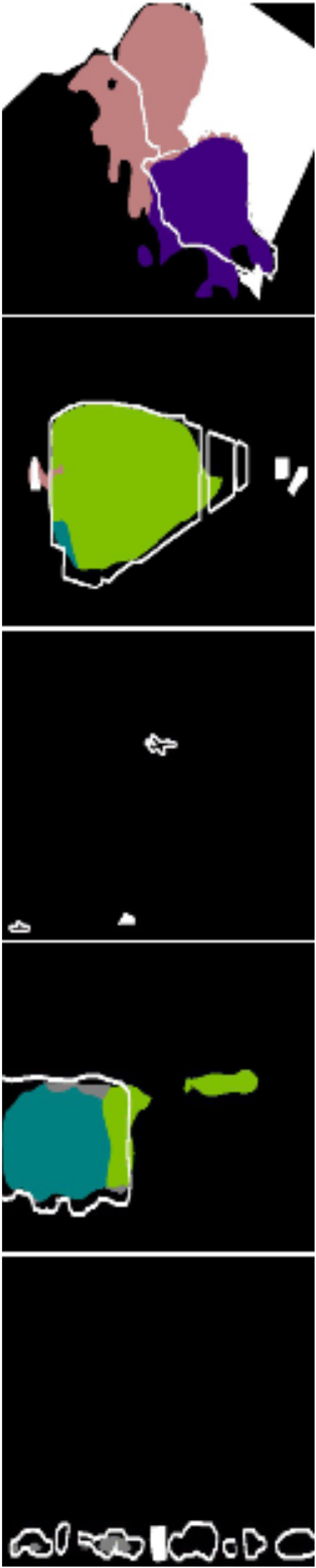}

}\subfloat[prune-2]{\includegraphics[width=0.155\columnwidth]{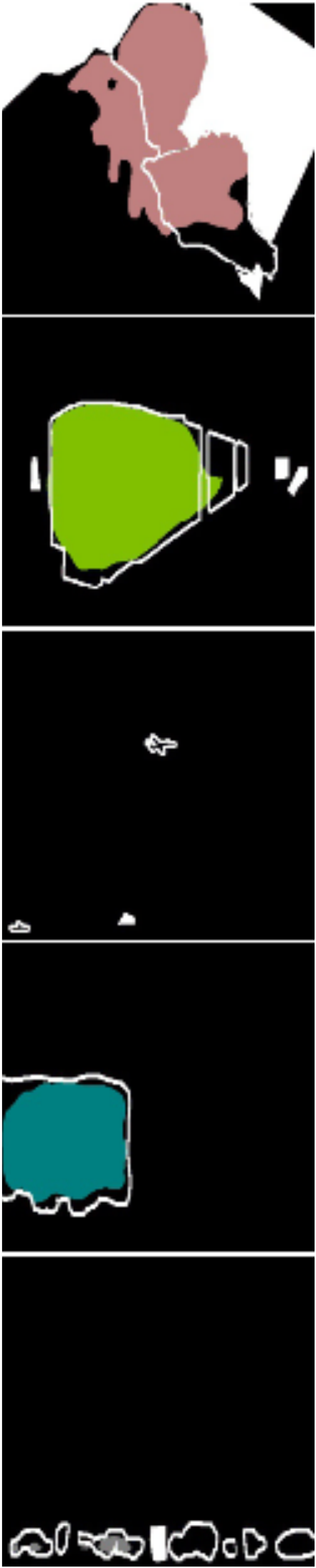}

}\subfloat[prune-1]{\includegraphics[width=0.155\columnwidth]{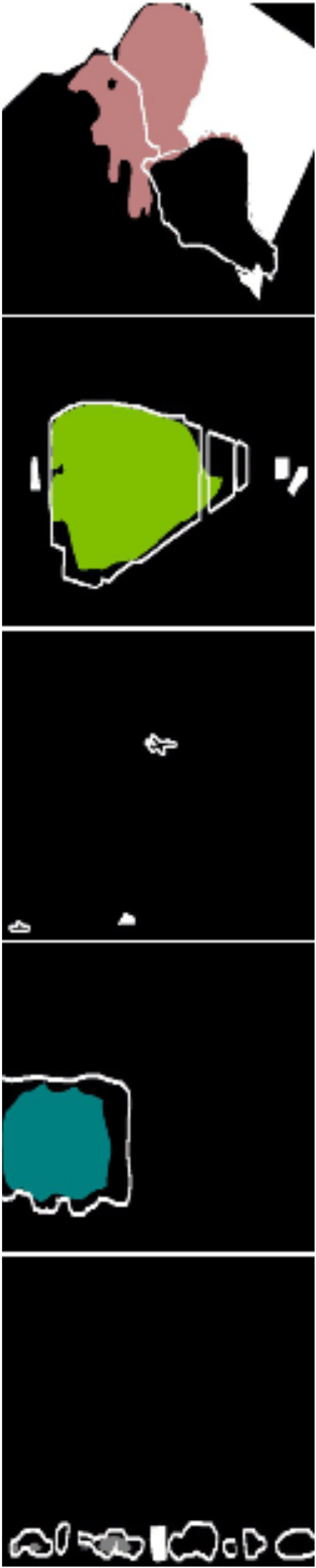}

}\subfloat[priming]{\includegraphics[width=0.155\columnwidth]{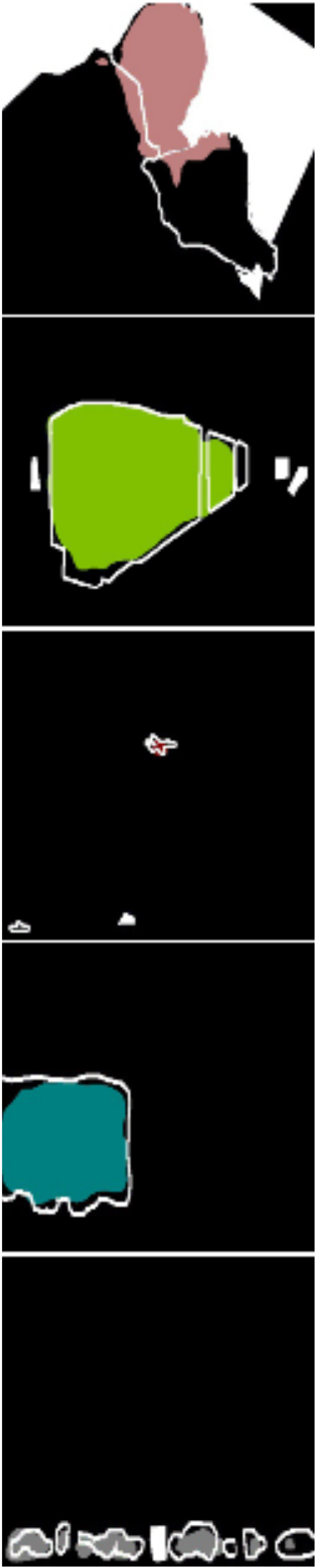}

}\caption{\label{fig:summary-segmentation}Comparing different methods of using
a cue to improve segmentation: From left to right: input image (with
cue overlayed), ground-truth (all classes), unprimed segmentation,
pruning type-2, pruning type-1, and priming. In each image, we aid
the segmentation network by adding a cue (e.g, ``plane''). White
regions are marked as ``don't care'' in the ground truth. }
\par\end{centering}
\end{figure}

As mentioned in Sec. \ref{par:Pruning}, we examine two types of pruning
to post-process segmentation results. One type removes image regions
which were wrongly labeled as the target class, replacing them with
background and the other increases the recall of previously missed
segmentation regions by removing all classes except the target class
and retaining pixels where the target class scored higher than the
background. The first type increases precision but cannot increase
recall. The second type increases recall but possibly hinders precision.
We found that both types results in a similar overall mean-IOU. Figure
\ref{fig:summary-segmentation} shows some examples where both types
of pruning result in segmentations inferior to the one resulting by
priming: post-processing can increase recall by lowering precision
(first row, column d) or increase precision by avoiding false-detections
(second and fourth row, column e), priming (column f) increases both
recall and precision. The second, and fourth rows missing parts of
the train/bus are recovered while removing false classes. The third
and fifth rows previously undetected small objects are now detected.
The person (first row) is segmented more accurately. 

\subparagraph{DeepLab.}

\begin{figure}
\includegraphics[width=1\columnwidth]{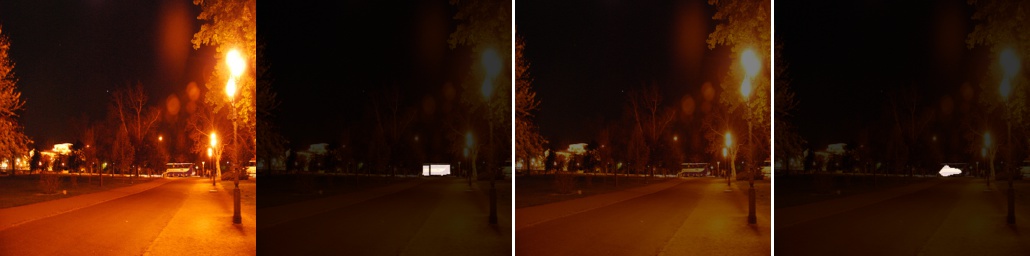}

\includegraphics[width=1\columnwidth]{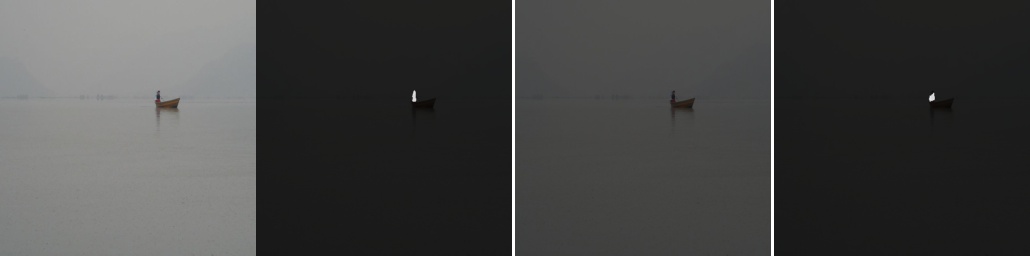}

\includegraphics[width=1\columnwidth]{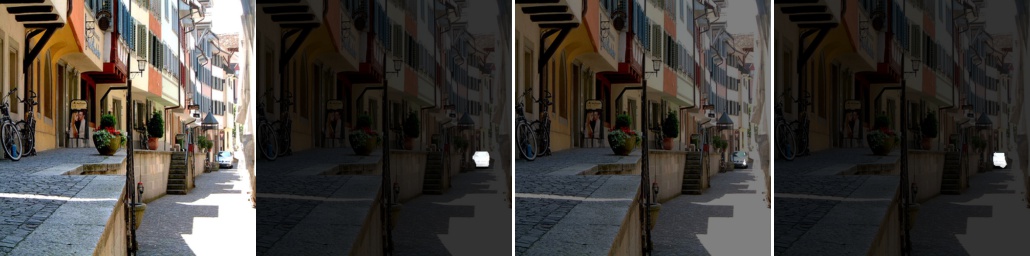}

\includegraphics[width=1\columnwidth]{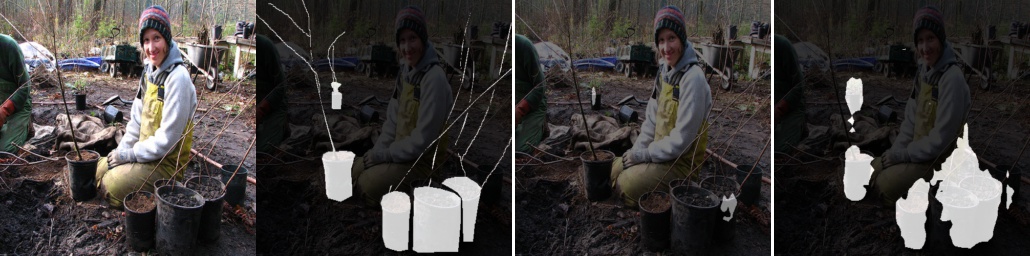}

\caption{\label{fig:Priming-a-network}Priming a network allows discovery of
small objects which are completely missed by the baseline, or ones
with uncommon appearance (last row). From left to right: input image,
ground-truth, baseline segmentation \cite{chen2016deeplab}, primed
network. }

\end{figure}

Next, we use the DeepLab \cite{chen2016deeplab} network for semantic-segmentation
with ResNet-101 \cite{he2016deep} as a base network. We do not employ
a CRF as post-processing. The mean-IOU of the baseline is 76.3\%.
Using Priming, increases this to 77.15\%. While in this case priming
does not improve as much as in the other cases we tested, we find
that it is especially effective at enabling the network to discover
small objects which were not previously segmented by the non-primed
version: the primed network discovers 57 objects which were not discovered
by the unprimed network, whereas the latter discovers only 3 which
were not discovered by the former. Fig. \ref{fig:Priming-a-network}
shows some representative examples of where priming was advantageous.
Note how the bus, person, (first three rows) are segmented by the
primed network (last column). We hypothesize that the priming process
helps increase the sensitivity of the network to features relevant
to the target object. The last row shows a successful segmentation
of potted plants with a rather atypical appearance. 

\subsubsection{Multi-Cue Object Detection}

We apply the CAT method to train priming on object detection as well.
For this experiment, we use the YOLOv2 method of \cite{redmon2016yolo9000}.
The base network we used is a port of the original network, known
as YOLOv2 544 \texttimes{} 544. Trained on the union of Pascal 2007
and 2012 datasets, it is reported by the authors to obtain 78.6\%
mAP on the test set of Pascal 2007. The implementation we use\footnote{\url{https://github.com/marvis/pytorch-yolo2}}
reaches a slightly lower 76.8\%, with a PyTorch port of the network
weights released by the authors. We use all the convolutional layers
of DarkNet (the base network of YOLOv2 ) to perform priming. We freeze
all network parameters of the original detection network and train
a priming network with the multi-cue training method for 25 epochs.
When using only pruning, performance on the test-set improves to 78.2\%
mAP. When we include priming as well, this goes up to 80.6\%,

\section{Conclusion}

We have presented a simple mechanism to prime neural networks, as
inspired by psychological top-down effects known to exist in human
observers. We have tested the proposed method on two tasks, namely
object detection and segmentation, using two methods for each task,
and comparing it to simple post-processing of the output. Our experiments
confirm that as is observed in humans, effective usage of a top-down
signal to modulate computations from early layers not only improves
robustness to noise but also facilitates better object detection and
segmentation, enabling detection of objects which are missed by the
baselines without compromising precision, notably so for small objects
and those having an atypical appearance. 

\bibliographystyle{plain}
\bibliography{egpaper_for_review}

\end{document}